\documentclass{article}
\usepackage{ijcai25}

\usepackage{times}
\usepackage{soul}
\usepackage{url}
\usepackage[hidelinks]{hyperref}
\usepackage[utf8]{inputenc}
\usepackage[small]{caption}
\usepackage{graphicx}
\usepackage{amsmath}
\usepackage{amsthm}
\usepackage{booktabs}
\usepackage{algorithm}
\usepackage[switch]{lineno}

\usepackage{bbm}
\usepackage{multirow}
\usepackage{multicol}
\usepackage{makecell}
\usepackage{color}
\usepackage{colortbl}
\definecolor{lightred}{rgb}{1, 0.8, 0.8}
\definecolor{lightblue}{rgb}{0.8, 0.9, 1}

\newcommand*\colourcheck[1]{%
  \expandafter\newcommand\csname #1check\endcsname{\textcolor{#1}{\ding{52}}}%
}
\colourcheck{blue}
\colourcheck{green}
\colourcheck{red}
\usepackage{pifont}
\usepackage{amsfonts}
\usepackage{algpseudocode}

\usepackage{natbib}

\usepackage[most]{tcolorbox}
\usepackage{listings}
\usepackage{float}

\tcbset{
  aibox/.style={
    width=474.18663pt,
    top=10pt,
    colback=white,
    colframe=black,
    colbacktitle=black,
    enhanced,
    center,
    attach boxed title to top left={yshift=-0.1in,xshift=0.15in},
    boxed title style={boxrule=0pt,colframe=white,},
  }
}
\newtcolorbox{AIbox}[2][]{aibox,title=#2,#1}

\title{Toward Reliable Scientific Hypothesis Generation: Evaluating Truthfulness and Hallucination in Large Language Models}

\author{
Guangzhi Xiong$^1$
\and
Eric Xie$^1$
\and
Corey Williams$^1$
\and
Myles Kim$^1$
\and
Amir Hassan Shariatmadari$^1$
\and
Sikun Guo$^1$
\and
Stefan Bekiranov$^1$
\And
Aidong Zhang$^1$
\\
\affiliations
$^1$University of Virginia\\
\emails
\{hhu4zu, jrg4wx, cmw6pa, mbt8hz, ahs5ce, qkm6sq, sb3de, aidong\}@virginia.edu
}

\begin{document}

\maketitle

\begin{abstract}

Large language models (LLMs) have shown significant potential in scientific disciplines such as biomedicine, particularly in hypothesis generation, where they can analyze vast literature, identify patterns, and suggest research directions. However, a key challenge lies in evaluating the truthfulness of generated hypotheses, as verifying their accuracy often requires substantial time and resources. Additionally, the hallucination problem in LLMs can lead to the generation of hypotheses that appear plausible but are ultimately incorrect, undermining their reliability. To facilitate the systematic study of these challenges, we introduce \textbf{TruthHypo}, a benchmark for assessing the capabilities of LLMs in generating truthful scientific hypotheses, and \textbf{KnowHD}, a knowledge-based hallucination detector to evaluate how well hypotheses are grounded in existing knowledge. Our results show that LLMs struggle to generate truthful hypotheses. By analyzing hallucinations in reasoning steps, we demonstrate that the groundedness scores provided by KnowHD serve as an effective metric for filtering truthful hypotheses from the diverse outputs of LLMs. Human evaluations further validate the utility of KnowHD in identifying truthful hypotheses and accelerating scientific discovery. Our data and source code are available at \url{https://github.com/Teddy-XiongGZ/TruthHypo}.

\end{abstract}

\section{Introduction}
Large language models (LLMs) have transformed the landscape of artificial intelligence, demonstrating remarkable capabilities across diverse applications, from natural language understanding to creative content generation \citep{karanikolas2023large,franceschelli2024creativity,raiaan2024review}. These models, trained on extensive corpora of text, demonstrate an ability to analyze, summarize, and generate human-like text, enabling advancements across diverse domains. Recently, there has been a growing interest in leveraging LLMs for scientific discovery \citep{zhong2023goal,yang2023large,kumar2023mycrunchgpt,liu2024conversational,baek2024researchagent,guo2024embracing}. Their capacity to process and synthesize vast amounts of scientific literature positions them as valuable tools in aiding researchers, particularly for tasks such as literature reviews, summarization, and even generating new hypotheses \citep{qi2023large,zhou2024hypothesis,m2024augmenting,wright2022generating,zeng2023meta,d2024marg,ifargan2025autonomous,yang2025large}.

One particularly promising application of LLMs is their use in scientific hypothesis generation, where they can assist in identifying promising research directions \citep{park2024can,si2024can,guo2025ideabench}. By analyzing extensive scientific literature, LLMs can uncover gaps in existing knowledge and propose novel hypotheses that may not be immediately apparent to human researchers. For instance, LLMs have been successfully applied to propose novel drug combinations for breast cancer treatment, some of which were later validated in laboratory experiments, showcasing their potential to accelerate biomedical discoveries \citep{abdel2024scientific}.

Despite these advancements, there are substantial challenges that limit the practical utility of LLMs in scientific hypothesis generation. A critical concern is the inability to evaluate the truthfulness of generated hypotheses. While LLMs can generate hypotheses that seem plausible, it remains uncertain whether these hypotheses are valid and grounded in existing knowledge or merely hallucinated and scientifically invalid. This issue is further exacerbated by the well-documented ``hallucination'' problem, where LLMs confidently produce information that is factually inaccurate or unsupported, posing challenges to their reliability in scientific contexts \citep{jin2024demystifying}. While current research has largely focused on improving the novelty and diversity of LLM-generated hypotheses, their truthfulness and grounding in established knowledge remain underexplored \citep{baek2024researchagent,hu2024nova,si2024can}.

\begin{figure*}[ht!]
    \centering
    \includegraphics[width=0.9\linewidth]{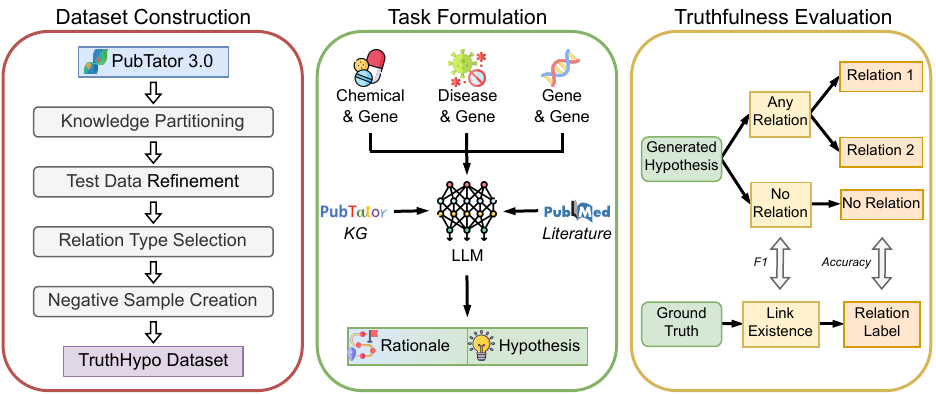}
    \caption{Overview of the TruthHypo benchmark, including dataset construction, task formulation, and truthfulness evaluation.}
    \label{fig:benchmark}
\end{figure*}

To address these challenges, we introduce TruthHypo, a comprehensive benchmark for evaluating the ability of LLMs to generate truthful scientific hypotheses, and KnowHD, a knowledge-based hallucination detection framework designed to assess the groundedness of these hypotheses. TruthHypo, built on a biomedical knowledge graph along with a domain-specific corpus, provides a controlled environment to evaluate how well LLM-generated hypotheses align with established scientific knowledge. KnowHD focuses on analyzing the reasoning processes of LLMs to identify hypotheses that are likely hallucinated or untruthful. Our findings reveal that LLMs face significant challenges in generating truthful hypotheses. By analyzing hallucinations in the reasoning processes behind generated hypotheses, we demonstrate that groundedness scores from KnowHD serve as an effective signal for identifying truthful hypotheses from the diverse outputs of LLMs. Human evaluations on open-ended hypothesis generation tasks further confirm the utility of KnowHD in identifying scientifically valid hypotheses.

Our main contributions are summarized as follows:
\begin{itemize}
\item We introduce TruthHypo, a comprehensive benchmark designed to evaluate the ability of LLMs to generate truthful scientific hypotheses.
\item We propose KnowHD, a knowledge-based hallucination detection framework that assesses the groundedness of LLM-generated hypotheses and identifies hallucinated claims by analyzing the rationale behind the hypothesis generation.
\item We provide an extensive analysis of existing LLMs on TruthHypo, highlighting their limitations and challenges in generating truthful hypotheses.
\item Our evaluation further reveals the connection between hallucination and truthfulness of generated hypotheses, showing the effectiveness of using KnowHD to select truthful and grounded hypotheses.
\end{itemize}
\section{Truthful Hypothesis Generation Benchmark} \label{sec:benchmark}

To systematically evaluate the ability of large language models (LLMs) to generate truthful scientific hypotheses, we introduce TruthHypo, a benchmark tailored for biomedical hypothesis generation. TruthHypo is designed to simulate real-world conditions by employing rigorous dataset construction, task formulation, and truthfulness evaluation metrics.
An overview of the dataset construction, task formulation, and evaluation framework is depicted in Figure \ref{fig:benchmark}.

\subsection{Dataset Construction}

The dataset for TruthHypo is derived from PubTator 3.0 \citep{wei2024pubtator}, a comprehensive biomedical knowledge graph that includes annotated relations (also called edges) extracted from scientific articles. To simulate the temporal progression of scientific discovery, we partitioned the graph into ``seen'' and ``unseen'' subsets based on the publication years of the corresponding articles. Relations in the ``seen'' subset were extracted from papers published before 2023, identified by PMIDs \(\leq 36600000\)\footnote{PMID is the unique identifier of the paper where the edge was extracted.}. The ``unseen'' subset, designed to represent new discoveries, comprises relations extracted from papers published after 2024, identified by PMIDs \(\geq 38200000\).

To ensure no overlap between the two subsets, we removed the edges in the unseen subset that shared head and tail entities with those in the seen subset. In addition, to maintain quality and validity, only relations discovered by multiple articles in the test data were retained. This filtering process guarantees that the unseen subset exclusively contains knowledge unavailable before 2024, simulating the conditions of future scientific research.

In building the dataset, we focused on three key relation types: ``Chemical \& Gene'', ``Disease \& Gene'', and ``Gene \& Gene''. These relation types were chosen for their complementary nature, detailed annotations, and potential for objective evaluation. To construct comprehensive classification tasks for evaluating different LLMs, we augment the dataset with negative test cases to assess whether LLMs tend to make false-positive predictions on entity pairs that lack a direct relationship in the existing knowledge base. The number of negative samples (labeled as ``no\_relation'') for each relation type is controlled to align with the average number of instances across other labels of the same relation type. The final dataset has 1209 instances for the ``Chemical \& Gene'' task, 268 instances for the ``Disease \& Gene'' task, and 547 instances for the ``Gene \& Gene'' task. A summary of the dataset statistics is presented in Table \ref{tab:benchmark_stats}.

\begin{table}[h]
    \centering
    \begin{tabular}{cccc}
    \toprule
        Task & Label & \# Instance \\
    \midrule
        Chemical \& Gene & \makecell{positive\_correlate\\negative\_correlate\\
        no\_relation} & \makecell{328\\478\\403} \\
        \midrule
        Disease \& Gene & \makecell{stimulate\\inhibit\\
        no\_relation} & \makecell{104\\75\\89} \\
        \midrule
        Gene \& Gene & \makecell{positive\_correlate\\negative\_correlate\\
        no\_relation} & \makecell{247\\118\\182} \\
    \bottomrule
    \end{tabular}
    \caption{Statistics of various tasks in the TruthHypo benchmark.}
    \label{tab:benchmark_stats}
\end{table}

\subsection{Task Formulation}

The TruthHypo benchmark includes three tasks, corresponding to the selected relation types: ``Chemical \& Gene'', ``Disease \& Gene'', and ``Gene \& Gene''. For each task, the input is a hypothesis generation query with two entities (see Figure \ref{fig:prompt_user_input} in Appendix \ref{sec:templates} for the template),  and the LLM is required to hypothesize the potential relationship between them based on available knowledge and reasoning.

To comprehensively assess LLM performance, we evaluate their ability to generate hypotheses under various knowledge augmentation settings. In the first setting, LLMs rely solely on their parametric knowledge -- information encoded in their parameters during pretraining on large corpora. This evaluates the model's intrinsic understanding and reasoning capabilities.

To enhance hypothesis generation, we introduce a second setting in which LLMs are augmented with structured knowledge from the ``seen'' knowledge graph. In this approach, key entities from the input are mapped to nodes in the graph, and multi-hop link chains connecting these nodes are explored. These chains, representing relevant relationships, are transformed into textual descriptions and provided as context for the model to use during hypothesis generation.

Another setting leverages information from biomedical literature using a retrieval-augmented generation (RAG) pipeline. Relevant documents are retrieved from the PubMed corpus\footnote{\url{https://pubmed.ncbi.nlm.nih.gov/}} using BM25 \citep{robertson2009probabilistic}. To maintain consistency with the knowledge graph's temporal split, only articles with PMIDs \(\leq 36600000\) are included in the retrieval. This simulates the process of generating hypotheses based on literature available at a given point in time.

Finally, we consider a combined setting, where both structured knowledge from the graph and unstructured information from retrieved literature are used to support hypothesis generation. This comprehensive approach provides a more holistic context, enabling the model to reason across both sources.
The LLM prompt templates we used to combine the external information with the original user instructions can be found in Figures \ref{fig:prompt_param}, \ref{fig:prompt_lit}, \ref{fig:prompt_kg}, and \ref{fig:prompt_kg_lit} (Appendix \ref{sec:templates}).

\subsection{Evaluation Metrics}

To evaluate the quality of generated scientific hypotheses, we employ a set of complementary metrics tailored to different aspects of hypothesis generation. These metrics assess the performance of LLMs in identifying valid connections between entities (link-level evaluation) and predicting specific relations (relation-level evaluation).

For link-level evaluation, we focus on precision, recall, and F1 score. Precision measures the proportion of correctly identified connections among all hypothesized connections, emphasizing the reduction of false positives. Recall evaluates the model’s ability to comprehensively identify all valid connections, capturing its sensitivity to true positives. The F1 score, as the harmonic mean of precision and recall, provides a balanced measure of performance, combining both the accuracy of predictions and the coverage of valid connections. These link-level metrics are critical for assessing the LLM’s ability to hypothesize plausible relationships between entities, regardless of the specific relation type.

For relation-level evaluation, we employ accuracy to measure how often the generated hypotheses match the correct relation labels in the ground truth. Accuracy captures the overall correctness of hypotheses by considering both the existence of a connection and the predicted relation type. While precision, recall, and F1 focus on identifying potential connections, accuracy provides a finer-grained assessment of the model’s capability to generate accurate relation labels.

By combining link-level and relation-level evaluations, the TruthHypo benchmark comprehensively measures the truthfulness of LLM-generated hypotheses, assessing the ability of LLMs to produce scientifically valid outputs.

\section{Knowledge-based Hallucination Detection} \label{sec:detection}

\begin{figure}[h]
    \centering
    \includegraphics[width=0.9\linewidth]{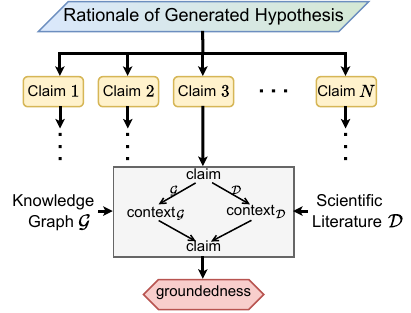}
    \caption{Overview of the KnowHD hallucination detection framework. Hypotheses are parsed into atomic claims, which are then evaluated for groundedness using a knowledge graph, scientific literature or both as knowledge sources.}
    \label{fig:hallucination}
\end{figure}

As discussed earlier, a critical concern regarding the truthfulness of LLM-generated hypotheses is the occurrence of hallucinations, where models generate plausible-sounding but unsupported claims. To address this, we introduce KnowHD, a knowledge-based hallucination detection framework that evaluates the groundedness of LLM-generated hypotheses by analyzing the rationale behind their generation. KnowHD operates using scientific literature, knowledge graphs, or a combination of both as the knowledge base. An overview of the framework is presented in Figure \ref{fig:hallucination}.

To evaluate groundedness, each hypothesis and its reasoning chain are first decomposed into a set of atomic claims. This step is critical because hypotheses often consist of compound reasoning steps, some of which may be supported by existing knowledge while others may not. Parsing these into atomic claims allows a more granular evaluation of groundedness and isolates unsupported components. This step is implemented by prompting LLMs with the template shown in Figure \ref{fig:prompt_claim} (Appendix \ref{sec:templates}).

When using scientific literature as the knowledge base, relevant documents for each atomic claim are retrieved from the PubMed corpus, limited to articles published before 2023 (PMID \(\leq 36600000\)). BM25 is employed to rank documents based on their relevance to the claim. To ensure computational efficiency and focus on the most relevant information, only the top-\(k\) documents are retained. The context retrieved from the literature corpus \(\mathcal{D}\) for a claim \(p\) is defined as:
\begin{equation}\label{eq:bm25}
\begin{aligned}
    \text{context}_{\mathcal{D}}(p)& =  \{d_1, d_2, \cdots, d_k | \\
    & d_i \in \mathcal{D}, \text{BM25}(p;d_i)\geq \tau, \text{rank}(d_i) \leq k \},
\end{aligned}
\end{equation}
where \(d_i\) represents a document in the corpus, \(\text{BM25}(p;d_i)\) is the relevance score assigned to the document for the claim \(p\). \(\tau\) is a threshold ensuring relevance, and \(\text{rank}(d_i)\) denotes the rank of \(d_i\) in the BM25-retrieved list. 

When using a knowledge graph \(\mathcal{G}\) as the knowledge base, the context for a claim is derived from the graph structure. For a claim \(p\), relevant knowledge is extracted as:
\begin{equation}
    \text{context}_{\mathcal{G}}(p) = \left\{(e_h, r, e_t) \in \mathcal{G} \left| \{e_h, e_t\} \subseteq \mathcal{V}(p) \right.\right\},
\end{equation}
where \((e_h, r, e_t)\) represents an edge in the knowledge graph with head entity \(e_h\), tail entity \(e_t\), and relation \(r\). The set \(\mathcal{V}(p)\) contains all entities mentioned in the claim \(p\). 

The groundedness of a claim is determined based on whether the given context information (\(\text{context}_{\mathcal{D}}\), \(\text{context}_{\mathcal{G}}\), or \(\text{context}_{\mathcal{D}}\cup\text{context}_{\mathcal{G}}\)) can fully support the claim, which is implemented by prompting LLMs to provide a judgment using the template in Figure \ref{fig:prompt_verification} (Appendix \ref{sec:templates}). If the concatenated context collectively entails the claim, it is considered grounded. The overall groundedness of a hypothesis \(h\) is computed as:
\begin{equation}
    \text{groudedness}(h) = \frac{1}{|\mathcal{C}(h)|} \sum_{p\in \mathcal{C}(h)} \mathbbm{1}[\text{context}(p) \models p],
\end{equation}
where \(\mathcal{C}(h)\) represents the set of atomic claims for hypothesis \(h\), and \(\mathbbm{1}[x \models y]\) returns 1 if \(x\) entails \(y\) and 0 otherwise. The \(\text{context}(p)\) can be \(\text{context}_{\mathcal{D}}(p)\), \(\text{context}_{\mathcal{G}}(p)\), or \(\text{context}_{\mathcal{D}}(p)\cup\text{context}_{\mathcal{G}}(p)\).

By offering both literature-based and graph-based contexts, KnowHD provides a robust framework for hallucination detection, offering flexibility to adapt to the available knowledge sources. This systematic evaluation of atomic claims enables a detailed assessment of the groundedness of hypotheses, identifying unsupported components and improving the reliability of LLM-generated outputs.
\section{Benchmark Analysis on TruthHypo}

\begin{table*}[htb!] \small
    \centering
    \begin{tabular}{cccccccccccccc}
    \toprule
    \multirow{2.5}{*}{\bf Knowledge} & \multirow{2.5}{*}{\bf LLM} & \multicolumn{2}{c}{\bf Chemical \& Gene} & \multicolumn{2}{c}{\bf Disease \& Gene} & \multicolumn{2}{c}{\bf Gene \& Gene} & \multicolumn{2}{c}{\bf Average} \\
    \cmidrule(lr){3-4} \cmidrule(lr){5-6} \cmidrule(lr){7-8} \cmidrule(lr){9-10}
    & & F1 & Acc & F1 & Acc & F1 & Acc & F1 & Acc \\
    \midrule
    \multirow{4}{*}{\makecell{Parametric\\\citep{wei2022chain}}} & Llama-3.1-8B & 80.16 & 42.43 & 79.37 & 41.04 & 79.19 & 46.07 & 66.90 & 43.23 \\
    & Llama-3.1-70B & 81.36 & 52.44 & 83.29 & 54.48 & 76.66 & 49.91 & 71.54 & 52.03 \\
    & GPT-4o-mini & 83.31 & 61.29 & 81.84 & 59.33 & 79.32 & 53.02 & 75.49 & 58.79 \\
    & GPT-4o & 80.74 & 66.17 & 75.38 & 54.85 & 71.56 & 55.58 & 73.17 & 61.81 \\
    \midrule
    \multirow{4}{*}{\makecell{Parametric + KG\\\citep{baek2024researchagent}}} & Llama-3.1-8B & 81.37 & 40.61 & 79.59 & 48.13 & 79.61 & 48.45 & 70.65 & 43.73 \\
    & Llama-3.1-70B & 87.85 & 62.86 & 67.62 & 52.24 & 78.29 & 58.14 & 79.10 & 60.18 \\
    & GPT-4o-mini & 86.42 & 57.65 & 74.17 & 55.60 & 81.65 & 62.34 & 79.40 & 58.65 \\
    & GPT-4o & 88.66 & 63.85 & 79.50 & 56.72 & 82.73 & 61.06 & 81.62 & 62.15 \\
    \midrule
    \multirow{4}{*}{\makecell{Parametric + Lit.\\\citep{lewis2020retrieval}}} & Llama-3.1-8B & 80.78 & 46.07 & 80.46 & 43.28 & 79.91 & 42.60 & 68.58 & 44.76 \\
    & Llama-3.1-70B & 82.56 & 56.74 & 84.16 & 52.99 & 79.18 & 51.55 & 73.37 & 54.84 \\
    & GPT-4o-mini & 85.28 & 59.80 & \bf 85.71 & 53.73 & 81.50 & 51.19 & 77.08 & 56.67 \\
    & GPT-4o & 79.52 & 65.92 & 75.84 & 55.97 & 64.69 & 51.92 & 71.84 & 60.82 \\
    \midrule
    \multirow{4}{*}{\makecell{Parametric + KG\\+ Literature}} & Llama-3.1-8B & 75.98 & 36.48 & 77.58 & 41.42 & 79.19 & 45.70 & 65.37 & 39.62 \\
    & Llama-3.1-70B & 84.80 & 59.31 & 77.64 & 56.34 & 81.24 & 55.76 & 77.37 & 57.95 \\
    & GPT-4o-mini & 88.34 & 60.96 & 84.47 & 58.21 & 84.17 & 58.50 & 81.42 & 59.93 \\
    & GPT-4o & \bf 89.71 & \bf 69.31 & 82.86 & \bf 62.31 & \bf 85.91 & \bf 63.99 & \bf 83.55 & \bf 66.95 \\
    \bottomrule
    \end{tabular}
    \caption{Performance comparison of different LLMs on the TruthHypo benchmark across various knowledge settings. The metrics reported are link-level F1 and relation-level accuracy (Acc) for each task (Chemical \& Gene, Disease \& Gene, Gene \& Gene), as well as their averages. ``Param.'' denotes parametric knowledge, while ``KG'' and ``Lit.'' refer to knowledge graphs and literature, respectively. All scores are percentages (\%).}
    \label{tab:main_table}
\end{table*}

\subsection{Experiment Settings}

To assess the ability of existing LLMs to generate truthful scientific hypotheses, we selected a diverse range of models varying in type and size. The Llama-3 family \citep{dubey2024llama} represents open-source LLMs, while the GPT-4 family \citep{achiam2023gpt} exemplifies proprietary models. From each family, we evaluated two LLMs of different sizes (Llama-3.1-8B \& Llama-3.1-70B, GPT-4o-mini \& GPT-4o) to investigate size-related differences in performance. All LLMs were trained on the knowledge available before 2024, preventing recall of the exact knowledge for hypothesis generation. More implementation details are in Appendix \ref{sec:implementation}.

The TruthHypo benchmark evaluates LLMs across four distinct settings: (1) parametric knowledge only, (2) parametric knowledge with knowledge graphs (KG), (3) parametric knowledge with literature (Lit.), and (4) parametric knowledge with both KG and literature. These settings allow us to explore the impact of external knowledge sources on hypothesis generation. The F1 and accuracy scores of different models are reported in this section. More detailed results on the precision and recall can be found in Appendix \ref{sec:additional}.

\subsection{Comparison of LLMs in Truthful Hypothesis Generation}

Table \ref{tab:main_table} presents the evaluation results for different LLMs and knowledge settings on TruthHypo. Across all tasks, the results indicate that most LLMs struggle to generate truthful scientific hypotheses, with only GPT-4o achieving mean accuracies exceeding 60\%. Additionally, we can observe that link-level F1 scores are higher than relation-level accuracy scores, which indicates that LLMs can identify potential connections between entities but often fail to accurately predict the specific relationships.

For models from the same family with different sizes, larger LLMs tend to generate scientific hypotheses more likely to be truthful. This can be attributed to two main factors. First, larger LLMs generally perform better because they can store and leverage more knowledge in their parameters, as shown by the results of parametric knowledge-only setting. Second, LLMs of different sizes have diverse capabilities to process external knowledge for hypothesis generation. For example, GPT-4o-mini shows a modest 1.14\% accuracy improvement when augmented with KG and literature, whereas GPT-4o achieves a more substantial 5.14\% increase under the same conditions. This suggests that larger LLMs can better utilize additional context to reason about truthful scientific hypotheses.
Similar trends are observed when comparing Llama-3.1-8B and Llama-3.1-70B. Interestingly, smaller models, such as Llama-3.1-8B, sometimes experience decreased performance when information from KG and literature is introduced. This degradation may stem from challenges in effectively integrating internal and external information, which can disrupt the model's reasoning processes.

Performance differences are also observed across the three relation types: ``Chemical \& Gene'', ``Disease \& Gene'' and ``Gene \& Gene''. Notably, all larger models, including GPT-4o, GPT-4o-mini, and Llama-3.1-70B, tend to perform better on ``Chemical \& Gene" tasks than on the other two types. This trend suggests that the ``Chemical \& Gene" task may be more aligned with the pre-trained knowledge or reasoning capabilities of these models. In contrast, the smaller Llama-3.1-8B shows a more inconsistent pattern, with performance varying across tasks and settings, likely reflecting its more limited parametric capacity and reasoning abilities.
These variations in performance across relation types may be attributed to differences in training data distributions or the complexity of the relation types themselves. The relatively stronger performance on the ``Chemical \& Gene" task highlights potential domain-specific biases or strengths in the LLMs, offering insights into their suitability for targeted applications in real-world scientific discovery.

\subsection{Hallucination Detection on LLM-generated Hypotheses}

To assess the groundedness of the generated hypotheses, we evaluated their rationales using KnowHD under various knowledge settings. KnowHD measures how well a hypothesis is supported by structured knowledge (KG), unstructured knowledge (literature), or both combined. The groundedness evaluation results for hypotheses generated by GPT-4o-mini are presented in Table \ref{tab:groundedness}.

\begin{table}[ht!] \small
    \centering
    \begin{tabular}{llcccc}
    \toprule
        \multirow{2.5}{*}{\bf Task} & \multirow{2.5}{*}{\bf Knowledge} & \multicolumn{3}{c}{\bf KnowHD} \\
        \cmidrule(lr){3-5} 
        & & KG & Lit. & KG + Lit. \\
    \midrule
        \multirow{4}{*}{\makecell[l]{Chemical\\\& Gene}} 
        & Parametric & 44.77 & 67.34 & 74.49 \\
        & + KG & 49.93 & 51.08 & 73.03 \\
        & + Lit. & 47.19 & 76.30 & 83.20 \\
        & + KG + Lit. & 50.57 & 65.25 & 78.90 \\
    \midrule
        \multirow{4}{*}{\makecell[l]{Disease\\\& Gene}} 
        & Parametric & 45.44 & 71.56 & 78.91 \\
        & + KG & 57.07 & 60.70 & 79.81 \\
        & + Lit. & 49.34 & 78.65 & 85.32 \\
        & + KG + Lit. & 51.11 & 75.26 & 86.68 \\
    \midrule
        \multirow{4}    {*}{\makecell[l]{Gene\\\& Gene}} 
        & Parametric & 42.94 & 67.81 & 76.16 \\
        & + KG & 58.07 & 56.41 & 79.64 \\
        & + Lit. & 44.49 & 76.43 & 84.48 \\
        & + KG + Lit. & 54.03 & 67.96 & 82.87 \\
    \bottomrule
    \end{tabular}
    \caption{KnownHD (KG, Lit., and KG + Lit.) groudedness scores of hypotheses generated by GPT-4o-mini under different knowledge settings. All scores are percentages (\%).}
    \label{tab:groundedness}
\end{table}

The results demonstrate distinct contributions of KG and literature to grounding hypotheses. For example, KnowHD with the literature as the support knowledge base can verify 76.30\% claims in the rationales of literature-augmented `Chemical \& Gene'' hypotheses. However, the hallucination detector can hardly verify the rationale generated based on adding KG information to parametric knowledge with only 51.08\% of the claims being grounded. Combining KG and literature yields the highest groundedness scores, effectively leveraging the complementary strengths of both sources to identify grounded claims and detect hallucinations.

\begin{figure*}[ht!]
    \centering
    \includegraphics[width=0.85\linewidth]{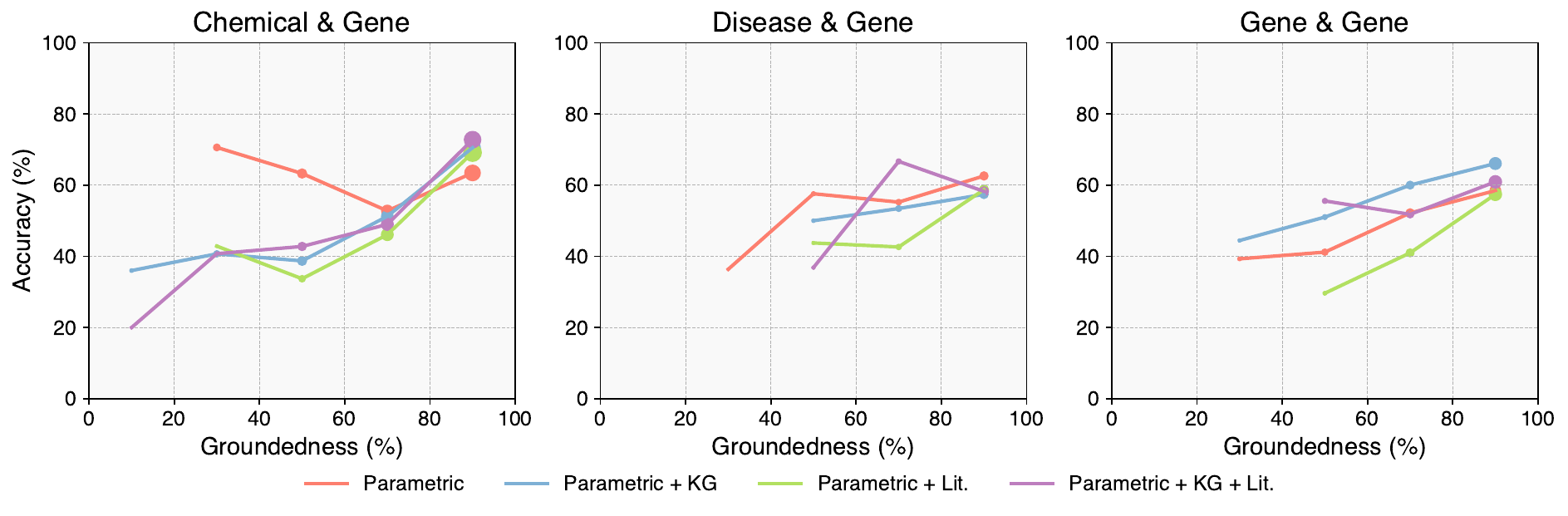}
    \caption{Mean accuracy corresponding to different levels of groundedness. Hypotheses are grouped based on their groundedness scores provided by KnowHD (KG + Literature). Only groups with no less than 10 hypotheses are shown in the plots. The dot size reflects the number of samples in each level of groundedness.}
    \label{fig:lines}
\end{figure*}

To further explore the relationship between hallucination and truthfulness, Figure \ref{fig:lines} examines mean accuracy as a function of groundedness scores. Hypotheses were grouped based on their groundedness scores, and the average accuracy for each group was calculated. The figure reveals a positive correlation between groundedness scores and hypothesis truthfulness. As groundedness scores increase, the likelihood of the hypothesis being truthful also increases. For example, GPT-4o-mini achieves a mean accuracy of 60.96\% on ``Chemical \& Gene" tasks under the combined KG + Literature setting, but this rises to 72.77\% for hypotheses with groundedness scores above 80\%. These findings underscore the potential of KnowHD to identify hypotheses with a higher probability of being truthful, particularly in contexts enriched with external knowledge.

\subsection{Improving Generation of Truthful Hypotheses with KnowHD}

To validate the utility of KnowHD on enhancing hypothesis generation, we prompted LLMs to generate five candidate hypotheses for each input and selected the one with the highest groundedness score as the final output. This approach was compared to two baselines: the greedy search method, where the hypothesis is generated using greedy next-token selection by the LLM, and the self-consistency method \citep{wang2022self}, which selects hypotheses based on majority voting across multiple predictions.

\begin{figure}[h!]
    \centering
    \includegraphics[width=0.85\linewidth]{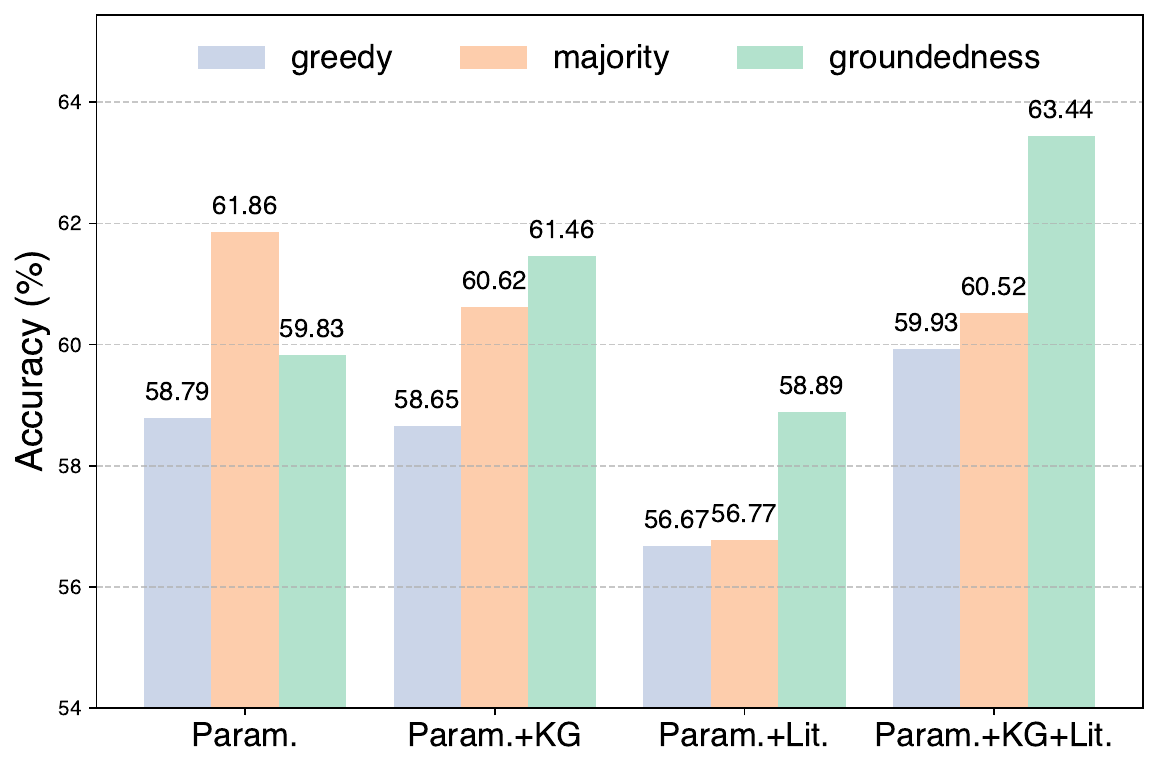}
    \caption{Accuracy improvements of GPT-4o-mini using KnowHD (KG + Lit.) groundedness scores for hypothesis selection. ``Param.'', ``KG'' and ``Lit.'' denote parametric knowledge, knowledge graphs, and literature, respectively.}
    \label{fig:improvement}
\end{figure}

As shown in Figure \ref{fig:improvement}, groundedness-based hypothesis selection generally outperforms both the greedy search and majority-voting methods across most knowledge settings. In the parametric knowledge-only setting, the majority-voting method achieves slightly higher accuracy (61.86\%) compared to groundedness-based selection (59.83\%). However, as external knowledge is introduced, groundedness-based selection demonstrates consistent improvements over both baselines. For example, in the combined parametric + KG + Literature setting, GPT-4o-mini achieves an average accuracy of 63.44\% when groundedness-based selection is used, approaching the performance of the larger GPT-4o model. 

These results highlight the effectiveness of groundedness scores in scenarios where external knowledge is incorporated, as they help identify hypotheses that are more likely to be truthful. By detecting hallucinations in reasoning steps and focusing on grounded hypotheses, KnowHD provides a robust mechanism for enhancing the reliability and truthfulness of LLM-generated scientific hypotheses.

\section{Human Study on Open-ended Tasks}

To further assess the generalizability of KnowHD's effectiveness in selecting truthful hypotheses, we conducted experiments on open-ended hypothesis generation tasks. These tasks were designed to evaluate whether KnowHD could reliably identify hypotheses with a higher likelihood of truthfulness across broader and less structured generation scenarios. For this analysis, we utilized the publicly available hypothesis generation dataset introduced by \citet{qi2024large}, which involves generating free-form hypotheses based on given background information.
We selected GPT-4o-mini as the tested LLM and enhanced its hypothesis generation process by incorporating external knowledge from scientific literature and knowledge graphs (KG). The model was prompted to generate five distinct scientific hypotheses for each input. These hypotheses were then evaluated by KnowHD, which assessed their groundedness based on their alignment with both structured (KG) and unstructured (literature) knowledge sources.

To analyze the relationship between groundedness scores and hypothesis truthfulness, we filtered generated hypotheses to create pairs with contrasting groundedness levels. For each input, we identified one hypothesis with the highest groundedness score and another with the lowest. We retained pairs where the higher groundedness score was at 30\% greater than the lower score. This filtering resulted in 54 pairs of hypotheses with significant differences in groundedness levels. To validate KnowHD’s effectiveness, we involved two domain experts to annotate each pair (80\% agreement), selecting the hypothesis they deemed more likely truthful based on the given information. Additionally, GPT-4o was prompted to analyze the same pairs and provide its judgment. Results of this annotation study, summarized in Table \ref{tab:open-ended}, report the selection ratio for each group, defined as the proportion of hypotheses in each group identified as more truthful.

\begin{table}\small
    \centering
    \begin{tabular}{l|ccccc}
    \toprule
        Group & Groudedness & \makecell{GPT-4o} & Human \\
    \midrule
        \makecell[l]{highly-grounded} & 83.94 & 61.11 & 59.26 \\
        \makecell[l]{lowly-grounded} & 40.61 & 38.89 & 40.74 \\
    \midrule
        \(p\)-value & 7.84 $\times 10^{-11}$  & 1.05 $\times 10^{-2}$ & 2.71 $\times 10^{-2}$  \\
    \bottomrule
    \end{tabular}
    \caption{Results of analysis on open-ended hypothesis generation tasks. ``GPT'' and ``Human'' denote the selection ratios by GPT-4o and human experts, respectively. All scores are percentages (\%). \(p\)-values were calculated using Wilcoxon signed-rank test and Z-test.}
    \label{tab:open-ended}
\end{table}

The results demonstrate a significant relationship between groundedness scores and the perceived truthfulness of hypotheses. Hypotheses with higher groundedness scores were consistently more likely to be selected as truthful by both human experts and GPT-4o, as indicated by the substantial differences in selection ratios. These findings highlight the utility of KnowHD in distinguishing truthful hypotheses, even in unstructured, open-ended generation tasks. By effectively leveraging groundedness as a criterion, KnowHD provides a robust mechanism for improving the reliability of LLM-generated hypotheses, reinforcing its potential for facilitating real-world scientific discovery processes.

\section{Related Work}

\subsection{Scientific Hypothesis Generation }

The use of LLMs for scientific hypothesis generation is a rapidly growing field, leveraging the ability of these models to process and synthesize vast amounts of scientific literature \citep{qi2023large,yang2023large,zhou2024hypothesis,ciucua2023harnessing,skarlinski2024language,radensky2024scideator,xiong2024scientific,guo2024embracing}. LLMs have been applied in identifying research gaps and generating novel hypotheses, with notable successes in areas such as drug discovery, where generated hypotheses have led to experimentally validated drug combinations \citep{abdel2024scientific}. 
Despite these advancements, most existing studies emphasize the novelty and diversity of hypotheses without addressing the critical aspect of truthfulness \citep{qi2024large,baek2024researchagent,wang2023scimon,hu2024nova,li2024chain}. The prevalent hallucination problem exacerbates this issue, as LLMs often generate hypotheses that appear plausible but lack factual support \citep{huang2023survey}. This gap motivates the development of TruthHypo, a benchmark explicitly designed to assess the ability of LLMs to generate truthful and grounded scientific hypotheses.

\subsection{Knowledge Graph Reasoning}

Knowledge graph reasoning involves inferring missing facts or relationships within a knowledge graph, with tasks such as link prediction, entity classification, and relation extraction being extensively studied \citep{nickel2015review,lin2015learning,ji2021survey,shu2024knowledge}. Traditional link prediction focuses on predicting edges between entities based on graph structure. These tasks primarily target structured graph completion, emphasizing pattern detection rather than creative reasoning \citep{zhang2018link,krenn2023forecasting,liu2023survey,wu2023dynamic,gu2024forecasting}.
TruthHypo introduces a novel benchmark that centers on LLM-driven scientific hypothesis generation, leveraging LLMs’ ability to flexibly integrate external knowledge through contextual inputs. Unlike static graph reasoning, TruthHypo evaluates how well LLMs generate grounded and truthful hypotheses. This shift highlights the growing role of LLMs in scientific discovery and bridges the gap between symbolic graph reasoning and natural language creativity.

\subsection{Retrieval-augmented Generation}

Retrieval-augmented generation (RAG) has emerged as a powerful approach for improving the factual accuracy and relevance of LLM outputs by integrating external knowledge during the generation process. This technique has been applied with literature retrieval, as demonstrated by \citep{lewis2020retrieval}, to dynamically incorporate up-to-date information into model outputs. Retrieval-augmented generation methods enhance the ability of LLMs to ground their outputs in external knowledge, making them particularly valuable in tasks requiring factual accuracy, such as scientific text generation \citep{lala2023paperqa,munikoti2023evaluating}.
In addition to literature retrieval, retrieval-augmented generation using knowledge graphs has gained attention for its potential to provide structured, domain-specific knowledge during text generation \citep{peng2024graph,ma2024think,wang2025knowledge}. TruthHypo builds on this paradigm by integrating both literature and knowledge graph retrieval to provide a robust evaluation of LLMs’ ability to generate truthful scientific hypotheses. This dual approach enables a comprehensive analysis of the role of external knowledge in mitigating hallucinations and ensuring the groundedness of generated hypotheses.

\section{Conclusion}
We presented TruthHypo, a benchmark for evaluating the ability of LLMs to generate truthful scientific hypotheses, and KnowHD, a framework for detecting hallucinations by assessing groundedness in reasoning. Through extensive evaluation, we highlighted the limitations of existing LLMs and demonstrated that selecting highly grounded hypotheses improves truthfulness. These contributions offer valuable insights for improving the reliability and utility of LLMs in scientific discovery.

\section*{Acknowledgements}
This work is supported in part by the US National Science Foundation under grants 2217071, 2213700, 2106913, 2008208, and NIH grant 1R01LM014012.

\bibliographystyle{named}
\bibliography{ijcai25}

\clearpage
\appendix
\section{Implementation Details} \label{sec:implementation}

For the retrieval of external knowledge from scientific literature, we implemented the information retrieval system by adopting the BM25 retriever \citep{robertson2009probabilistic} for processed PubMed chunks provided by the MedRAG toolkit \citep{xiong2024benchmarking,xiong2024improving}. BM25 (Best Matching 25) is a probabilistic retrieval model that ranks documents based on term frequency, document length normalization, and the specificity of terms through inverse document frequency (IDF). We selected BM25 as our text retriever because it is particularly effective for the biomedical domain, where dense retrievers often struggle to encode the nuanced semantics of biomedical terms such as gene names \citep{luo2022improving}. BM25's reliance on exact term matching with statistical weighting makes it well-suited for capturing term-specific relevance in structured biomedical text. In our experiments, \(\tau\) in Equation \eqref{eq:bm25} is set as 0.0. The number of retrieved documents is set as \(k=32\) for hypothesis generation, and \(k=8\) for claim verification.

To identify biomedical entities in a given claim, we used a two-step process. In the first step, we prompted LLMs to extract the entity mentions directly from the claim (Figure \ref{fig:prompt_entity}). This step focused on identifying relevant biomedical terms, such as gene names, proteins, or diseases, without additional processing or complex workflows. The extracted entity mentions were then used in the second step, where each mention was matched to its unified representation in the PubTator 3.0 knowledge graph. This matching was implemented using a BM25 retriever. For constructing the BM25 index, each piece of text, or ``chunk'', was designed by concatenating all possible mentions of a given entity stored in PubTator 3.0. By leveraging BM25's ranking capabilities, we retrieved the most relevant chunk corresponding to each entity mention, ensuring accurate alignment with PubTator's unified entities. 

\section{Computational Cost}

Table \ref{tab:token} shows the number of all tokens used in experiments for Table \ref{tab:main_table}. It shows that the additional knowledge from either the knowledge graph (KG) or literature (Lit.) will significantly increase the number of input tokens. In particular, the literature brings more tokens than KG, as the knowledge in KG is always structured and summarized. While input lengths vary across different settings, output lengths are relatively stable, a consistent pattern shown in different LLMs.

\begin{table}[h!] \small
    \centering
    \begin{tabular}{cccccccc}
    \toprule
        \multirow{2.5}{*}{\bf LLM} & \multirow{2.5}{*}{\bf Type} & \multicolumn{4}{c}{\bf Setting} \\
        \cmidrule(lr){3-6}
        & & Param. & +KG & +Lit. & +KG+Lit. \\
    \midrule
        \multirow{2}{*}{\makecell{Llama\\-3.1-8B}} & Input & 295.8k & 1.7M & 25.8M & 27.2M \\
         & Output & 811.1k & 1.0M & 782.5k & 1.2M \\
    \midrule
        \multirow{2}{*}{\makecell{Llama\\-3.1-70B}} & Input & 295.8k & 1.7M & 25.8M & 27.2M \\
         & Output & 813.7k & 881.2k & 777.6k & 767.8k \\
    \midrule
        \multirow{2}{*}{\makecell{GPT-4o\\-mini}} & Input & 295.8k & 1.7M & 25.8M & 27.2M \\
         & Output & 751.6k & 684.0k & 787.2k & 707.1k \\
    \midrule
        \multirow{2}{*}{\makecell{GPT-4o}} & Input & 295.8k & 1.7M & 25.8M & 27.2M \\
         & Output & 909.9k & 839.1k & 891.5k & 875.3k \\
    \bottomrule
    \end{tabular}
    \caption{Summary of \#tokens used for all experiments in Table \ref{tab:main_table}.}
    \label{tab:token}
\end{table}

\section{Additional Quantitative Results on TruthHypo} \label{sec:additional}

Table \ref{tab:main_table} presents the F1 score of various LLMs on the TruthHypo benchmark, evaluating their ability to identify the existence of a new relation given current knowledge. To provide a more granular analysis, Table \ref{tab:prec_recall} breaks down the results into precision and recall for different tasks, offering insights into the strengths and weaknesses of each model and knowledge augmentation setting.

From the results in Table \ref{tab:prec_recall}, we observe that smaller LLMs, such as Llama-3.1-8B, tend to achieve higher recall scores across all tasks, along with relatively lower precision. This indicates that while these models can generate a comprehensive set of hypotheses, they are prone to a high false positive rate, which could pose challenges in real-world applications, such as scientific hypothesis generation, where precision is often critical. High false positive rates could result in wasted time and resources when pursuing hypotheses that are unlikely to hold upon experimental validation.

Given that validating new biomedical hypotheses often requires months or even years of research, ensuring high precision in hypothesis generation is of paramount importance. Among the tested models, GPT-4o with external knowledge from the literature achieved the highest precision across all tasks, demonstrating its ability to generate hypotheses with fewer false positives. However, this precision came at the expense of lower recall, especially when compared to GPT-4o with knowledge augmentations from both literature and knowledge graphs (KG). This trade-off highlights the importance of balancing precision and recall based on the specific requirements of a given application.

When comparing different knowledge settings, we found that the improvements provided by external knowledge sources varied across tasks and models. For example, knowledge graph (KG) information significantly enhanced the precision of all LLMs on tasks involving ``Disease \& Gene'' and ``Gene \& Gene'' relations, but it did not notably improve the precision of GPT-4o on the ``Chemical \& Gene'' task. In contrast, the literature knowledge augmentation slightly improved the precision of all LLMs except GPT-4o-mini. Interestingly, the setting that combined both knowledge sources provided a more balanced precision improvement, offering a middle ground between the individual benefits of KG and literature-based augmentations.

Additionally, Table \ref{tab:prec_recall} reveals that larger models such as GPT-4o consistently outperformed smaller models in precision, regardless of the knowledge setting, reflecting their ability to integrate complex external information effectively. This highlights the potential of larger models to better utilize structured and unstructured knowledge sources for hypothesis generation. However, smaller models, with their higher recall, may still serve as useful tools for exploratory or broad hypothesis generation tasks where exhaustive coverage is prioritized over precision.

Overall, the analysis demonstrates that the choice of LLM and knowledge augmentation strategy should be guided by the specific trade-offs between precision and recall that align with the requirements of the downstream task. For biomedical applications, where precision is often paramount, leveraging models like GPT-4o with literature-based augmentations appears to be the most effective approach.

\begin{table*}[h!]
    \centering
    \begin{tabular}{cccccccccccccc}
    \toprule
    \multirow{2.5}{*}{\bf Knowledge} & \multirow{2.5}{*}{\bf LLM} & \multicolumn{2}{c}{\bf Chemical \& Gene} & \multicolumn{2}{c}{\bf Disease \& Gene} & \multicolumn{2}{c}{\bf Gene \& Gene} & \multicolumn{2}{c}{\bf Average} \\
    \cmidrule(lr){3-4} \cmidrule(lr){5-6} \cmidrule(lr){7-8} \cmidrule(lr){9-10}
    & & Prec & Recall & Prec & Recall & Prec & Recall & Prec & Recall \\
    \midrule
    \multirow{4}{*}{\makecell{Parametric\\\cite{wei2022chain}}} & Llama-3.1-8B & 67.57 & \bf 98.51 & 66.79 & \bf 97.77 & 66.92 & \bf 96.99 & 67.29 & \bf 98.00 \\
    & Llama-3.1-70B & 74.69 & 89.33 & 75.23 & 93.30 & 73.13 & 80.55 & 74.37 & 87.48 \\
    & GPT-4o-mini & 83.00 & 83.62 & 79.47 & 84.36 & 75.50 & 83.56 & 80.37 & 83.70 \\
    & GPT-4o & 90.59 & 72.83 & 82.67 & 69.27 & 83.27 & 62.74 & 87.60 & 69.63 \\
    \midrule
    \multirow{4}{*}{\makecell{Parametric + KG\\\cite{baek2024researchagent}}} & Llama-3.1-8B  & 71.42 & 94.54 & 74.04 & 86.03 & 72.16 & 88.77 & 71.93 & 91.85 \\
    & Llama-3.1-70B & 90.63 & 85.24 & 93.14 & 53.07 & 88.58 & 70.14 & 90.34 & 76.89 \\
    & GPT-4o-mini & 86.37 & 86.48 & 91.06 & 62.57 & \bf 92.39 & 73.15 & 88.27 & 79.70 \\
    & GPT-4o & 86.27 & 91.19 & \bf 91.30 & 70.39 & 87.96 & 78.08 & 87.21 & 84.89 \\
    \midrule
    \multirow{4}{*}{\makecell{Parametric + Lit.\\\cite{lewis2020retrieval}}} & Llama-3.1-8B  & 68.82 & 97.77 & 68.36 & \bf 97.77 & 68.49 & 95.89 & 68.67 & 97.26 \\
    & Llama-3.1-70B & 74.92 & 91.94 & 75.56 & 94.97 & 74.58 & 84.38 & 74.92 & 90.30 \\
    & GPT-4o-mini & 78.18 & 93.80 & 80.10 & 92.18 & 74.94 & 89.32 & 77.55 & 92.37 \\
    & GPT-4o & \bf 92.73 & 69.60 & 83.78 & 69.27 & 89.37 & 50.68 & \bf 90.62 & 64.44 \\
    \midrule
    \multirow{4}{*}{\makecell{Parametric + KG\\+ Literature}} & Llama-3.1-8B & 68.21 & 85.73 & 70.64 & 86.03 & 69.96 & 91.23 & 69.01 & 87.26 \\
    & Llama-3.1-70B & 84.13 & 85.48 & 87.41 & 69.83 & 80.05 & 82.47 & 83.33 & 82.59 \\
    & GPT-4o-mini & 82.61 & 94.91 & 82.45 & 86.59 & 83.16 & 85.21 & 82.73 & 91.19 \\
    & GPT-4o & 86.61 & 93.05 & 84.80 & 81.01 & 87.50 & 84.38 & 86.61 & 89.11 \\
    \bottomrule
    \end{tabular}
    \caption{Performance comparison of different LLMs on the TruthHypo benchmark across various knowledge settings, with precision and recall as the evaluation metrics. ``Prec'' denotes the link-level precision, while ``Recall'' represents the link-level recall.}
    \label{tab:prec_recall}
\end{table*}

To further understand the limitations of hypothesis generation with high groundedness scores, we conducted an in-depth analysis of the error patterns. We identified two representative types of errors: (1) cases where the LLM incorrectly infers that there is no association between the given entities, despite supporting evidence; and (2) cases where the model simply echoes or paraphrases the provided context without engaging in substantive reasoning or hypothesis formation. These findings highlight the need for more robust mechanisms to ensure both accurate association detection and genuine reasoning in hypothesis generation, enhancing the interpretability and trustworthiness of the overall system \citep{doshi2017towards,loh2022application,miller2023explainable,sinha2024colidr,sinha2024self}.

\section{Prompt Templates for LLMs in Experiments} \label{sec:templates}

Figure \ref{fig:prompt_user_input} shows the template we used to construct a hypothesis generation query given two different entities.
The prompt templates for the use of LLMs in the ``Parametric'', ``Parametric + KG'', ``Parametric + Lit.'', and ``Parametric + KG + Lit.'' settings are presented in Figures \ref{fig:prompt_param}, \ref{fig:prompt_kg}, \ref{fig:prompt_lit}, \ref{fig:prompt_kg_lit}, respectively. These templates were designed to guide the LLMs in effectively leveraging various sources of knowledge while maintaining a pre-determined structure in the model output to facilitate consistent parsing and downstream analysis.

Figure \ref{fig:prompt_claim} shows the template for LLMs to extract scientific claims from the entire rationale. For the identification of biomedical entities and the use of LLMs for claim verification, we employed the templates shown in Figures \ref{fig:prompt_entity} and \ref{fig:prompt_verification}, respectively. The entity identification templates (Figure \ref{fig:prompt_entity}) were crafted to enable the LLMs to extract precise mentions of biomedical entities such as genes or diseases from textual claims. These prompts were carefully designed to minimize ambiguity, ensuring that entities sharing the same mention could be properly distinguished using their unique IDs.

\begin{figure*}[h]
\begin{AIbox}{Prompt template for constructing user input with given entities}
Can we hypothesize the potential relation between \verb|{{entity type 1}}| \verb|{{entity name 1}}| \verb|({{entity ID 1}}|) and \verb|{{entity type 2}}| \verb|{{entity name 2}}| \verb|({{entity ID 2}}|)? The final hypothesis can be one of [\verb|{{relation label 1}}|, \verb|{{relation label 2}}|, `no\_relation'].
\end{AIbox}
\caption{Prompt template for constructing user input with given entities.}
\label{fig:prompt_user_input}
\end{figure*}

\begin{figure*}[h]
\begin{AIbox}{Prompt template for hypothesis generation with parametric knowledge only}
You are a scientist. Your task is to generate a scientific hypothesis following given instructions.\\

\#\#\# User Input\\
\verb|{{input}}|\\

Your output must include two sections:\\
1. **\#\#\# Step-by-step Reasoning**:\\
- Think step-by-step to derive the hypothesis.\\

2. **\#\#\# Structured Output**:\\
- Present your proposed hypothesis in the following JSON format:\\
\texttt{```}json\\
  \{\\
      ``proposed\_hypothesis'': ``Statement of the proposed hypothesis''\\
  \}\\
  \texttt{```}\\
\end{AIbox}
\caption{Prompt template for hypothesis generation with parametric knowledge only.}
\label{fig:prompt_param}
\end{figure*}

\begin{figure*}[h]
\begin{AIbox}{Prompt template for hypothesis generation with knowledge from parameters and KG}
You are a scientist. Your task is to generate a scientific hypothesis following given instructions.\\

\#\#\# Relevant Knowledge\\
\verb|{{knowledge}}|\\

\#\#\# User Input\\
\verb|{{input}}|\\

Your output must include two sections:\\
1. **\#\#\# Step-by-step Reasoning**:\\
- Think step-by-step to derive the hypothesis.\\

2. **\#\#\# Structured Output**:\\
- Present your proposed hypothesis in the following JSON format:\\
\texttt{```}json\\
  \{\\
      ``proposed\_hypothesis'': ``Statement of the proposed hypothesis''\\
  \}\\
  \texttt{```}\\
\end{AIbox}
\caption{Prompt template for hypothesis generation with knowledge from parameters and knowledge graphs (KGs).}
\label{fig:prompt_kg}
\end{figure*}

\begin{figure*}[h]
\begin{AIbox}{Prompt template for hypothesis generation with knowledge from parameters and literature}
You are a scientist. Your task is to generate a scientific hypothesis following given instructions.\\

\#\#\# Relevant  Documents\\
\verb|{{document}}|\\

\#\#\# User Input\\
\verb|{{input}}|\\

Your output must include two sections:\\
1. **\#\#\# Step-by-step Reasoning**:\\
- Think step-by-step to derive the hypothesis.\\

2. **\#\#\# Structured Output**:\\
- Present your proposed hypothesis in the following JSON format:\\
\texttt{```}json\\
  \{\\
      ``proposed\_hypothesis'': ``Statement of the proposed hypothesis''\\
  \}\\
  \texttt{```}\\
\end{AIbox}
\caption{Prompt template for hypothesis generation with knowledge from parameters and literature.}
\label{fig:prompt_lit}
\end{figure*}

\begin{figure*}[h]
\begin{AIbox}{Prompt template for hypothesis generation with knowledge from parameters and KG and literature}
You are a scientist. Your task is to generate a scientific hypothesis following given instructions.\\

\#\#\# Relevant  Documents\\
\verb|{{document}}|\\

\#\#\# Relevant Knowledge\\
\verb|{{knowledge}}|\\

\#\#\# User Input\\
\verb|{{input}}|\\

Your output must include two sections:\\
1. **\#\#\# Step-by-step Reasoning**:\\
- Think step-by-step to derive the hypothesis.\\

2. **\#\#\# Structured Output**:\\
- Present your proposed hypothesis in the following JSON format:\\
\texttt{```}json\\
  \{\\
      ``proposed\_hypothesis'': ``Statement of the proposed hypothesis''\\
  \}\\
  \texttt{```}\\
\end{AIbox}
\caption{Prompt template for hypothesis generation with knowledge from parameters, KG, and literature.}
\label{fig:prompt_kg_lit}
\end{figure*}

\begin{figure*}[h]
\begin{AIbox}{Prompt template for claim identification}
\#\#\# Statement\\
\verb|{{statement}}|\\

Summarize the statement as a list of claims which will be further verified by external resources. Output the summarized claims in the JSON format: \texttt{```}json\{``claims'': [``claim1'', ...]\}\texttt{```}\\
\end{AIbox}
\caption{Prompt template for claim identification.}
\label{fig:prompt_claim}
\end{figure*}

\begin{figure*}[h]
\begin{AIbox}{Prompt template for entity recognition}
\#\#\# Background\\
\verb|{{background}}|\\

Extract key entities from the background statement that will be used to search for relevant information in an external knowledge graph. Each entity should be extracted as ``entity\_type (e.g., Disease/Chemical/Gene/Mutation) entity\_name (entity\_id if presented)''. Output the extracted entities in the JSON format: \texttt{```}json\{``entities'': [``entity1'', ...]\}\texttt{```}\\
\end{AIbox}
\caption{Prompt template for entity recognition.}
\label{fig:prompt_entity}
\end{figure*}

\begin{figure*}[h]
\begin{AIbox}{Prompt template for claim verification}
You are a scientist. Your task is to verify if the relevant documents and knowledge (if applicable) can support the given claim.\\

\#\#\# Relevant  Documents\\
\verb|{{document}}|\\

\#\#\# Relevant Knowledge\\
\verb|{{knowledge}}|\\

\#\#\# Claim\\
\verb|{{claim}}|\\

Judge if the given information supports the claim. Output \{``groundedness'': 1\} if the materials support the claim else \{``groundedness'': 0\}.\\
\end{AIbox}
\caption{Prompt template for claim verification.}
\label{fig:prompt_verification}
\end{figure*}

\end{document}